\documentclass{article} 
\usepackage{nips14submit_e,times}
\usepackage{hyperref}
\usepackage{url}

\usepackage{graphicx}
\usepackage{transparent}
\usepackage{subcaption}
\usepackage{amsmath}
\usepackage{amssymb}
\usepackage{cite}

\title{Flexible Deep Neural Network Processing}

\footnotesize{\author{
Hokchhay Tann \hspace{0.4in} Soheil Hashemi \hspace{0.4in} Sherief Reda\\
School of Engineering \\
Brown University \\
Providence, RI \\
\footnotesize{\texttt{\{hokchhay\_tann, soheil\_hashemi, sherief\_reda\}@brown.edu}} \\
}}

\nipsfinalcopy 

\begin{document}

\maketitle

\begin{abstract}
The recent success of Deep Neural Networks (DNNs) has drastically improved the state of the art for many application domains. 
While achieving high accuracy performance, deploying state-of-the-art DNNs is a challenge since they typically require billions of expensive arithmetic computations. In addition, DNNs are typically deployed in ensemble to boost accuracy performance, which further exacerbates the system requirements. This computational overhead is an issue for many platforms, e.g. data centers and embedded systems, with tight latency and energy budgets. In this article, we introduce flexible DNNs ensemble processing technique, which achieves large reduction in average inference latency while incurring small to negligible accuracy drop. Our technique is flexible in that it allows for dynamic adaptation between quality of results (QoR) and execution runtime. We demonstrate the effectiveness of the technique on AlexNet and ResNet-50 using the ImageNet dataset. This technique can also easily handle other types of networks.


\end{abstract}

\section{Introduction}

With increased availability of computational power in recent years, deep neural networks (DNNs) have shown to generate state-of-the-art performance in many complex machine learning and computer vision problems. This performance leap has led to widespread use of DNNs across industries, from data centers to embedded devices, and explosive increase in the academic research in the area. Currently, DNNs power all major image and voice recognition systems including Apple and Google with many further applications emerging. Other industries such as health care and finance have also started adopting this technology. Recently, DNNs have shown to outperform radiologists in spotting pneumonia \cite{chexnet}.

Such industry-wide adoption of DNNs has proliferated their deployment to a wide range of systems, from battery-operated embedded platforms to massive data centers, which presents significant runtime challenges. In addition, while DNNs dramatically improve accuracy performance of many machine learning problems, developing new network architectures and learning methods which improves state-of-the-art is a very tedious process. For this reason, a popular method to boost inference accuracy performance is to apply ensemble learning, where each input is not evaluated by a single model but using multiple independently trained DNNs. This further exacerbates deployment challenges on time-pressured systems such as user-oriented web services, which require strict latency enforcement \cite{GOOGLE:TPU}. For systems with constrained computational resources such as autonomous vehicles \cite{NVIDIA:END2END} and embedded platforms \cite{CODES,DATE:17,DAC:17}, low-latency and energy inference is desirable for both runtime performance and lifetime of the systems.

Foundational works targeting general resource-constrained deployment of machine learning models introduce \textit{flexible} computation methodologies, where partial results can be accepted in exchange for reduction in allocations of costly resources such as time and memory \cite{horvitz0,horvitz1}. Since the inference output may have a time-dependent utility, waiting for increasingly accurate results could have a net negative impact. The costs of delayed actions and increase in computations may outweighs the benefits of a more accurate inference. By adopting flexible computing approach, large computation and latency reductions can be made while incurring small drop in quality of results (QoR). The constraint of this flexibility is that the overall utility of the systems must not decrease.

In the deployment of DNN ensembles, each additional model evaluation linearly increases the overall latency and required computations of the systems. 
For platforms with real-time delay/energy constraints, processing every input using all the models in the ensemble may not always be efficient or even possible. To address this issue, we propose flexible ensemble processing as a form of flexible computation, where input data is only evaluated using additional DNN model based on a {\it metareasoner}. The metareasoner computes the likelihood of increase utility with additional model evaluations and decides whether to continue or output the current results. With this flexible computation model, we achieve large reduction in average runtime per input while still maintaining most of benefits from ensemble learning.

\textbf{Our contribution: } We introduce flexible ensemble processing methodology offering large runtime and energy reduction with small inference accuracy drop compared to normal ensemble execution. We demonstrate our technique on two well-known DNNs, namely AlexNet and ResNet-50.

\section{Related Works} \label{prev_work}

For many machine learning problems, in order to reduce generalization error, a simple and effective technique is often to deploy an ensemble of models. For problems with small models and datasets, techniques such as Bootstrap Aggregating (bagging) \cite{bagging} are often used. On the other hand, larger models such as DNNs normally reach a large variety of solutions simply by using different initializations, which takes away the need for special data partitioning \cite{dlbook}. For this reason, DNNs ensembles can be formed using a single model architecture by independently training them with different initializations. Zheng \cite{zheng} demonstrates this in practical application, where an ensembles of DNNs show better prediction capability of software-reliability than a single model. The effectiveness of model ensembles is explained by Dietterich \cite{dietterich} as the consequence of three fundamental reasons: statistics, computations, and representations.

While DNN ensembles dramatically improve the QoR for many machine learning problems, deploying such large models incurs longer latency and requires massive resource. Horvitz and Rutledge \cite{horvitz1} demonstrate examples where the increases in inference latency and system resource allocations could nullify any additional utility gains from the extra computations. In fact, continuing to wait for a higher QoR may decrease the overall utility of the systems. Addressing this problem, Horvitz \cite{horvitz0} introduces flexible computation methodologies, where knowledge of model utility is used to control the trade offs between additional computations and acting with partial results. In this work, we propose a methodology to enable flexible DNN ensemble processing in order to minimize the overall systems latency while trading off small inference accuracy loss.

\section{Methodology}
\label{methodology}

In this section, we discuss the DNNs ensemble method and our proposed flexible processing technique targeted at lowering inference latency and computational demand.

\subsection{Ensemble of Deep Neural Networks}

Deploying an ensemble of DNNs has been proven to be a simple and reliable method to boost the inference accuracy. After independently training multiple DNNs of the same architecture, each input is then evaluated using all of the networks. The final output of the model is computed by combining the outputs from all the DNNs in the ensemble. The combination process can be a weighted or unweighted voting. For this study, we use unweighted outputs averaging. For instance, suppose there are $N$ networks in the ensemble, and for each input, they produce output logit vectors $\textbf{z}_i$, $i\in[1,N]$. Then the final output vector is computed as $\hat{\textbf{z}}=\frac{1}{N}\sum_{i=1}^N\textbf{z}_i$. In the case of classification, the predicted class can simply be the maximum element in $\hat{\textbf{z}}$. 

\begin{figure*}[t]
\vspace{-0.1in}
   \begin{center}
    \includegraphics[scale=0.55,trim={1.5cm 7cm 1.5cm 8cm},clip]{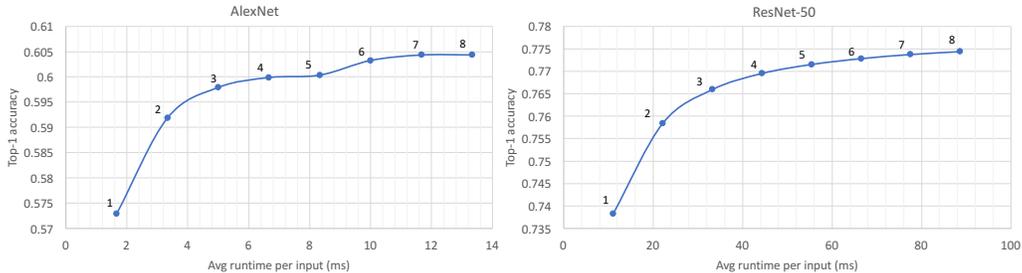}
    \caption{Inference accuracy versus average runtime per input for AlexNet and ResNet-50 for DNN ensembles on ImageNet validation set. Each data label shows the number of networks in the ensemble. Runtime results are based on a system with a Nvidia Titan Xp GPU.}
    \label{fig:acc}
    \end{center}
    \vspace{-0.2in}
\end{figure*}

Figure \ref{fig:acc} shows the top-1 accuracy on ImageNet validation set versus average runtime per input for AlexNet and ResNet-50 for ensembles with different number of networks. As shown, for both DNNs, the inference accuracy improves for ensembles with increasing number of DNNs. Based on Figure \ref{fig:acc}, this improvement seems logarithmic to the number of models in the ensembles. The accuracy saturates after ensemble of 7 models for AlexNet while it continues to improve with 8 models for ResNet-50. Although the accuracy improvement diminishes and saturates with larger ensembles, it still represent a significant boost compared to single network performance. 


Next, we discuss our proposed methodology to cut down the average runtime, and hence the energy per input.

\subsection{Flexible Deep Neural Network Ensemble Processing} \label{flexible}
The goal of ensemble execution is to improve overall inference accuracy of the model. However, executing all networks in the ensemble for every input incurs high costs for valuable resources such as time and memory. As shown in Figure \ref{fig:acc}, increasing ensemble size linearly increases the processing latency while it logarithmically improves performance accuracy. In our proposed methodology, our aim is to preserve this accuracy improvement, while significantly reducing the average latency per input. We do this by introducing a metareasoner which determines when it is unnecessary to execute additional networks. For instance, for an ensemble of 8 DNNs, instead of running each input through all 8 DNNs, we first process the input using one DNN. Then, based on the metareasoner, we only evaluate using additional DNNs as necessary.

In similar fashion to Horvitz \cite{horvitz1}, the metareasoner reasons about the probability of utility increase with additional latency and resource allocations. The utility can be thought of as the value of additional computations performed. A net positive value of computation increases the utility of the systems. In our case, we define the net value of computation as positive for situations where it is highly probable that the current inference output is incorrect and that additional model evaluation may change that. This decision model introduces processing flexibility in that every input is evaluated using only a number of models in the ensembles considered optimal. 

\begin{figure}[t]
   \begin{center}
    \includegraphics[scale=0.55,trim={1.5cm 8cm 1.5cm 8cm},clip]{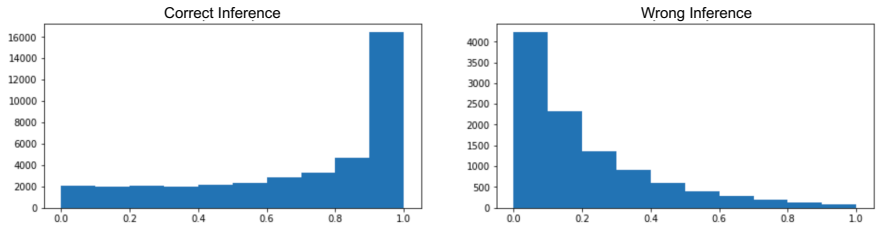}
    \caption{Score Margins histograms for correct and wrong top-1 inference for AlexNet. The x-axis shows the score margin, and the y-axis shows the number of samples in each score margin bin.}
    \label{fig:scoremargin}
    \end{center}
    \vspace{-0.1in}
\end{figure}

In order to compute the probability of a net positive value of computation, we analyze the score margins of the current inference output. {\it Score margin} is defined as the absolute difference between the top two scores (logits) in the DNN output. For instance, suppose $\textbf{P}$ is a pre- or post-softmax output vector for a DNN, the score margin is defined as
\begin{align*}
	SM &=|m0 - m1|\\
    \textnormal{where }m0 &= \max(\textbf{P})\\
    \textnormal{and }m1 &= \max(\{\textbf{P}\}\textnormal{\textbackslash} m0).
\end{align*}
It is observed that there exists a strong correlation between the top two score margins and the prediction accuracy \cite{park}. This observation is also observed for our models as shown in Figure \ref{fig:scoremargin}. Here, we show two histogram plots of the score margins for when the inference is correct and when it is wrong based on the true data labels for a single DNN. This figure is generated using a random subset of 50000 images from the training data. With this correlation, the net value of computation can be estimated by comparing the score margin to a set threshold.


\begin{figure}[t]
   \begin{center}
    \includegraphics[scale=0.7,trim={3.7cm 8.5cm 3.7cm 7.5cm},clip]{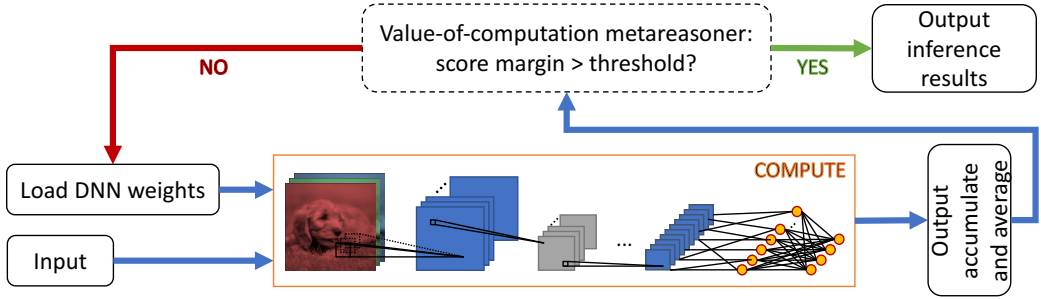}
    \caption{Execution flow for flexible DNN ensemble processing.}
    \label{fig:flexible}
    \end{center}
    \vspace{-0.1in}
\end{figure}

Figure \ref{fig:flexible} illustrates our flexible DNN ensemble processing. Once an input is evaluated though a DNN, the score margin is computed and passed to the metareasoner, where it is compared against a set threshold. We use post-softmax output, so our score margin is always in the range $[0,1]$. If the margin is higher than the threshold, it highly probable that the current is already correct, and additional model processing will likely produce a net negative value of computation. For this reason, the processing is halted and current prediction is output as the final inference result. Otherwise, we execute the additional DNN model and average the prediction. This process is repeated until, we finish executing all the DNN in the ensemble. Since the number of DNNs in the ensemble between each execution changes, we set different thresholds for each ensemble size. We empirically choose the thresholds based on the training set data so as to minimize latency and QoR loss. Our objective function is $M=\alpha\cdot R + (1-\alpha)\cdot E$, where $R$ is the inference latency and $E$ is the relative increase in inference error compared to normal ensemble execution. $\alpha$ determines the relative importance of error rate increase and latency improvement. Since our score margin is in the range $[0,1]$, we perform a grid search for each ensemble size over different threshold values and output the value that minimizes $M$.

\section{Experimental Results}
In this section, we evaluate the runtime benefit and accuracy impact from our methodology.

\subsection{Experimental Setup}
For our experiments, we measure deployment runtime of the flexible ensemble execution using a system with Intel Core i7 4790K CPU and a Nvidia Titan Xp GPU. This setup allows us to analyze the runtime benefit on smaller scale systems, where DNNs in an ensemble is executed serially. Note that the saving reported on this system should also be observed on smaller embedded platforms, where only one DNN can be executed at a time. Our accuracy results are based on the ImageNet 2012 datasets. We employ two well-known DNN architectures namely AlexNet \cite{Krizhevsky} and ResNet-50 \cite{resnet}. All of our experiments are based on Caffe \cite{caffe}.

\textit{Decision model:} The score margin threshold choices directly affect the inference latency versus accuracy trade off. Lower threshold values would mean each input is likely to get by less number of DNNs, which results in shorter average latency. However, this would also mean that the accuracy is lower. Thus, selecting optimal threshold is crucial. Since this work is application agnostic, we achieve optimal threshold by setting $\alpha=0.5$ in Section \ref{flexible}.

\subsection{Results and Discussions}
\begin{figure*}[t]
\vspace{-0.1in}
   \begin{center}
    \includegraphics[scale=0.55,trim={1.4cm 6.8cm 1.4cm 6.5cm},clip]{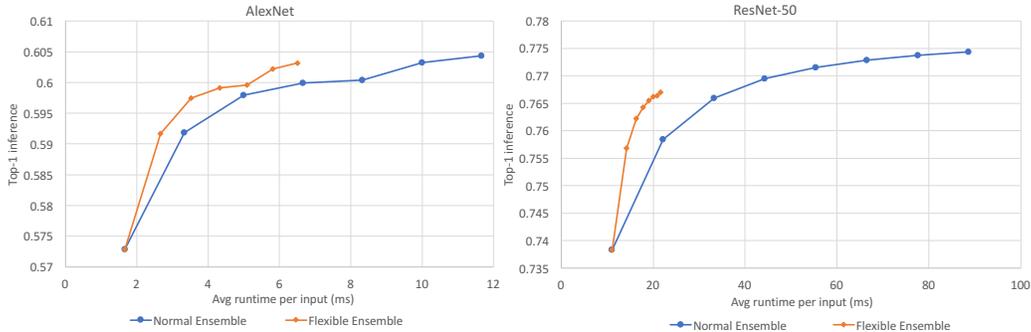}
    \caption{Inference accuracy versus average runtime per input for AlexNet and ResNet-50 for normal and flexible ensemble execution. Runtime results are based on a system with a Nvidia Titan Xp GPU.}
    \label{fig:flexibleresult}
    \end{center}
    \vspace{-0.2in}
\end{figure*}

Figure \ref{fig:flexibleresult} shows the inference latency and accuracy for two execution modes, normal and our proposed flexible ensemble executions. For AlexNet, Figure \ref{fig:acc} shows that ensemble with 8 DNNs achieves no accuracy gain compared to that with 7 DNNs. For this reason, we only show results for ensemble with up to 7 DNNs in Figure \ref{fig:flexibleresult}. For both of the DNNs presented, flexible ensemble processing retains majority of the inference accuracy of normal DNN ensembles while offering large reduction in average latency. For instance, in AlexNet case, for an ensemble of 7 DNNs, the average runtimes for normal and flexible executions are 11.67 ms and 6.51 ms respectively while the accuracies are 0.6043 and 0.6032 respectively. This is close to 2$\times$ latency improvement with 0.1\% inference accuracy drop. Using our methodology, this drop can be traded off with the latency improvement by adjusting the score margin thresholds. In the extreme case, where no accuracy drop is tolerable, we can set the score margin thresholds very high, which is equivalent to normal ensemble execution and would not result in any relative accuracy loss.

Bounded-resource inference has long been a pressing issue in machine learning problems. Flexible computing introduces alternative inference strategies, where QoR is gracefully traded off for benefits in lower computational costs \cite{horvitz0,horvitz1}. Toward this goal, we presented a flexible execution methodology that lessens DNN ensemble computation and latency overheads while still maintaining much of the inference accuracy. This technique allows large degree of freedoms for inference accuracy versus latency trade off, and it can be readily combined with other types of approximations. In addition, this approach can be easily extended to handle other types of neural networks or other kinds of machine learning models.


\subsubsection*{Acknowledgments}

This work is supported in part by NSF grant number 1420864. We would like to thank NVIDIA Corporation for generous GPU donation.

\end{document}